\documentclass[11pt,letterpaper]{article}
\usepackage{emnlp2017}
\usepackage{times}
\usepackage{latexsym}
\usepackage[subtle]{savetrees}
\usepackage{array}
\usepackage{amsmath}
\usepackage{amsfonts}
\usepackage{multirow}
\usepackage{booktabs}
\usepackage{subcaption}
\usepackage{caption}
\usepackage{graphicx}
\usepackage[colorinlistoftodos]{todonotes}

\usepackage{etoolbox}

\emnlpfinalcopy

\def\emnlppaperid{***}

\title{Efficient Attention using a Fixed-Size Memory Representation}

\author{
    Denny Britz\thanks{\hspace{0.1 cm}Equal Contribution. Author order alphabetical.} \and Melody Y. Guan\footnotemark[1] \and Minh{-}Thang Luong \\ Google Brain \\ \texttt{dennybritz,melodyguan,thangluong@google.com}
}

\date{}

\begin{document}

\maketitle

\begin{abstract} 
The standard content-based attention mechanism typically used in sequence-to-sequence models is computationally expensive as it requires the comparison of large encoder and decoder states at each time step. In this work, we propose an alternative attention mechanism based on a fixed size memory representation that is more efficient. Our technique predicts a compact set of $K$ attention contexts during encoding and lets the decoder compute an efficient lookup that does not need to consult the memory. We show that our approach performs on-par with the standard attention mechanism while yielding inference speedups of 20\% for real-world translation tasks and more for tasks with longer sequences. By visualizing attention scores we demonstrate that our models learn distinct, meaningful alignments.
\end{abstract}


\section{Introduction}
\label{sec:introduction}

Sequence-to-sequence models ~\cite{Sutskever:2014, Cho:2014} have achieved state of the art results across a wide variety of tasks, including Neural Machine Translation (NMT) ~\cite{Bahdanau:2014, Wu:2016}, text summarization ~\cite{Rush:2015, Nallapati:2016}, speech recognition ~\cite{Chan:2015, Jan:2016}, image captioning ~\cite{Xu:2015}, and conversational modeling ~\cite{Vinyals:2015, Li:2015}.

The most popular approaches are based on an encoder-decoder architecture consisting of two recurrent neural networks (RNNs) and an attention mechanism that aligns target to source tokens ~\cite{Bahdanau:2014, Luong:2015}. The typical attention mechanism used in these architectures computes a new attention context at each decoding step based on the current state of the decoder. Intuitively, this corresponds to looking at the source sequence after the output of every single target token. 


Inspired by how humans process sentences, we believe it may be unnecessary to look back at the entire original source sequence at each step.\footnote{Eye-tracking and keystroke logging data from human translators show that translators generally do not reread previously translated source text words when producing target text ~\cite{carl2011taxonomy}.}  We thus propose an alternative attention mechanism (section \ref{sec:model}) that leads to smaller computational time complexity. Our method predicts $K$ attention context vectors while reading the source, and learns to use a weighted average of these vectors at each step of decoding. Thus, we avoid looking back at the source sequence once it has been encoded. We show (section \ref{sec:experiments}) that this speeds up inference while performing on-par with the standard mechanism on both toy and real-world WMT translation datasets. We also show that our mechanism leads to larger speedups as sequences get longer. Finally, by visualizing the attention scores (section \ref{sec:visualizing_attention}), we verify that the proposed technique learns meaningful alignments, and that different attention context vectors specialize on different parts of the source.

\section{Background}
\label{sec:background}
\subsection{Sequence-to-Sequence Model with Attention}

Our models are based on an encoder-decoder architecture with attention mechanism ~\cite{Bahdanau:2014,Luong:2015}. An encoder function takes as input a sequence of source tokens $\mathbf{x} = (x_1, ..., x_m)$ and produces a sequence of states $\mathbf{s} = (s_1, ..., s_m)$ .The decoder is an RNN that predicts the probability of a target sequence $\mathbf{y} = (y_1, ..., y_T \mid \mathbf{s})$. The probability of each target token $y_i \in \{1, ... ,|V|\}$ is predicted based on the recurrent state in the decoder RNN, $h_i$, the previous words, $y_{<i}$, and a context vector $c_i$. The context vector $c_i$, also referred to as the attention vector, is calculated as a weighted average of the source states.

\begin{align}
c_i & = \sum_{j}{\alpha_{ij} s_j} \\
{\alpha}_{i} & = \text{softmax}(f_{att}(h_i, \mathbf{s}))
\end{align}

Here, $f_{att}(h_i, \mathbf{s})$ is an attention function that calculates an unnormalized alignment score between the encoder state $s_j$ and the decoder state $h_i$. Variants of $f_{att}$ used in ~\citet{Bahdanau:2014} and ~\citet{Luong:2015} are:
\[
    f_{att}(h_i, s_j)= 
\begin{cases}
    v_a^T \text{tanh}(W_a[h_i, s_j]),& \emph{Bahdanau} \\
    h_i^TW_as_j & \emph{Luong} 
\end{cases}
\]
where $W_a$ and $v_a$ are model parameters learned to predict alignment. Let $|S|$ and $|T|$ denote the lengths of the source and target sequences respectively and $D$ denoate the state size of the encoder and decoder RNN. Such content-based attention mechanisms result in inference times of $O(D^2|S||T|)$\footnote{An exception is the dot-attention from ~\citet{Luong:2015}, which is $O(D|S||T|)$, which we discuss further in Section 3.}, as each context vector depends on the current decoder state $h_i$ and all encoder states, and requires an $O(D^2)$ matrix multiplication.

The decoder outputs a distribution over a vocabulary of fixed-size $|V|$:
\begin{align}P(y_i \vert y_{<i}, \mathbf{x}) = \text{softmax}(W[s_i; c_i] + b)\end{align}
The model is trained end-to-end by minimizing the negative log likelihood of the target words using stochastic gradient descent.

\section{Memory-Based Attention Model}
\label{sec:model}

\begin{figure*}[ht]
    \centering
    \includegraphics[width=0.75\textwidth]{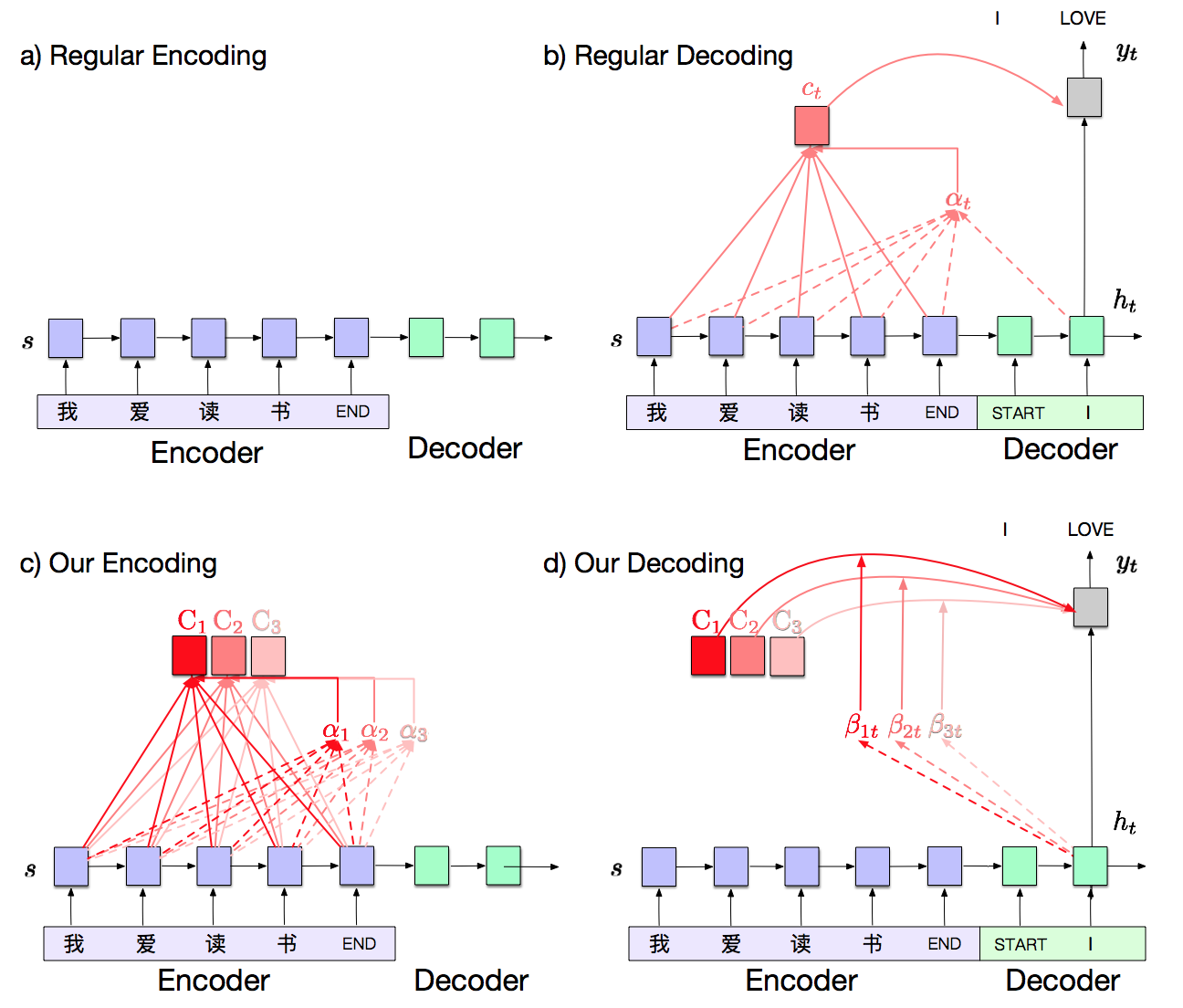}
    \caption{Memory Attention model architecture. $K$ attention vectors are predicted during encoding, and a linear combination is chosen during decoding. In our example, $K=3$.}
    \label{fig:model}
\end{figure*}

Our proposed model is shown in Figure \ref{fig:model}. During encoding, we compute an attention matrix $C \in \mathbb{R}^{K \times D}$, where $K$ is the number of attention vectors and a hyperparameter of our method, and $D$ is the dimensionality of the top-most encoder state. This matrix is computed by predicting a score vector $\alpha_t \in \mathbb{R}^K$ at each encoding time step $t$. $C$ is then a linear combination of the encoder states, weighted by $\alpha_t$:
\begin{align}
    C_k & = \sum_{t=0}^{|S|}{\alpha_{tk} s_t} \\
    \alpha_t & = \text{softmax}(W_\alpha s_t) ,
\end{align}
where $W_{\alpha}$ is a parameter matrix in $\mathbb{R}^{K\times D}$. 

The computational time complexity for this operation is $O(KD|S|)$. One can think of C as compact fixed-length memory that the decoder will perform attention over. In contrast, standard approaches use a variable-length set of encoder states for attention. At each decoding step, we similarly predict $K$ scores $\beta \in \mathbb{R}^K$. The final attention context $c$ is a linear combination of the rows in $C$ weighted by the scores.  Intuitively, each decoder step predicts how important each of the $K$ attention vectors is.

\begin{align}
    c & = \sum_{i=0}^{K}{\beta_i C_i} \\
    \beta & = \text{softmax}(W_\beta h)
\end{align}
Here, $h$ is the current state of the decoder, and $W_\beta$ is a learned parameter matrix. Note that we do not access the encoder states at each decoder step. We simply take a linear combination of the attention matrix $C$ pre-computed during encoding - a much cheaper operation that is independent of the length of the source sequence. The time complexity of this computation is $O(KD|T|)$ as multiplication with the $K$ attention matrices needs to happen at each decoding step. 

Summing $O(KD|S|)$ from encoding and $O(KD|T|)$ from decoding, we have a total linear computational complexity of $O(KD(|S| + |T|)$. As $D$ is typically very large, $512$ or $1024$ units in most applications, we expect our model to be faster than the standard attention mechanism running in $O(D^2|S||T|)$. For long sequences (as in summarization, where |S| is large), we also expect our model to be faster than the cheaper dot-based attention mechanism, which needs $O(D|S||T|)$ computation time and requires encoder and decoder states sizes to match.

We also experimented with using a sigmoid function instead of the softmax to score the encoder and decoder attention scores, resulting in 4 possible combinations. We call this choice the \emph{scoring function}. A softmax scoring function calculates normalized scores, while the sigmoid scoring function results in unnormalized scores that can be understood as gates.



\subsection{Model Interpretations}

Our memory-based attention model can be understood intuitively in two ways. We can interpret it as "predicting" the set of attention contexts produced by a standard attention mechanism during encoding. To see this, assume we set $K \approx |T|$. In this case, we predict all $|T|$ attention contexts during the encoding stage and learn to choose the right one during decoding. This is cheaper than computing contexts one-by-one based on the decoder and encoder content. In fact, we could enforce this objective by first training a regular attention model and adding a regularization term to force the memory matrix $C$ to be close to the $T\times D$ vectors computed by the standard attention. We leave it to future work to explore such an objective.

Alternatively, we can interpret our mechanism as first predicting a compact $K \times D$ memory matrix, a representation of the source sequence, and then performing \textit{location-based} attention on the memory by picking which row of the matrix to attend to. Standard location-based attention mechanism, by contrast, predicts a location in the source sequence to focus on ~\cite{Luong:2015, Xu:2015}.

\subsection{Position Encodings (PE)}

In the above formulation, the predictions of attention contexts are symmetric. That is, $C_i$ is not forced to be different from $C_{j\neq i}$. While we would hope for the model to learn to generate distinct attention contexts, we now present an extension that pushes the model into this direction. We add \emph{position encodings} to the score matrix that forces the first few context vector $C_1, C_2, ...$ to focus on the beginning of the sequence and the last few vectors $...,C_{K-1}, C_K$ to focus on the end (thereby encouraging in-between vectors to focus on the middle).

Explicitly, we multiply the score vector $\alpha$ with position encodings $l_s\in \mathbb{R}^{K}$:
\begin{align}
    C^{PE} & = \sum_{s=0}^{|S|}{\alpha^{PE} h_s} \\
    \alpha^{PE}_s & = \text{softmax}(W_\alpha h_s \circ l_s)
\end{align}

\begin{figure}
    \includegraphics[trim={2cm 0 0 0},clip,width=0.5\textwidth]{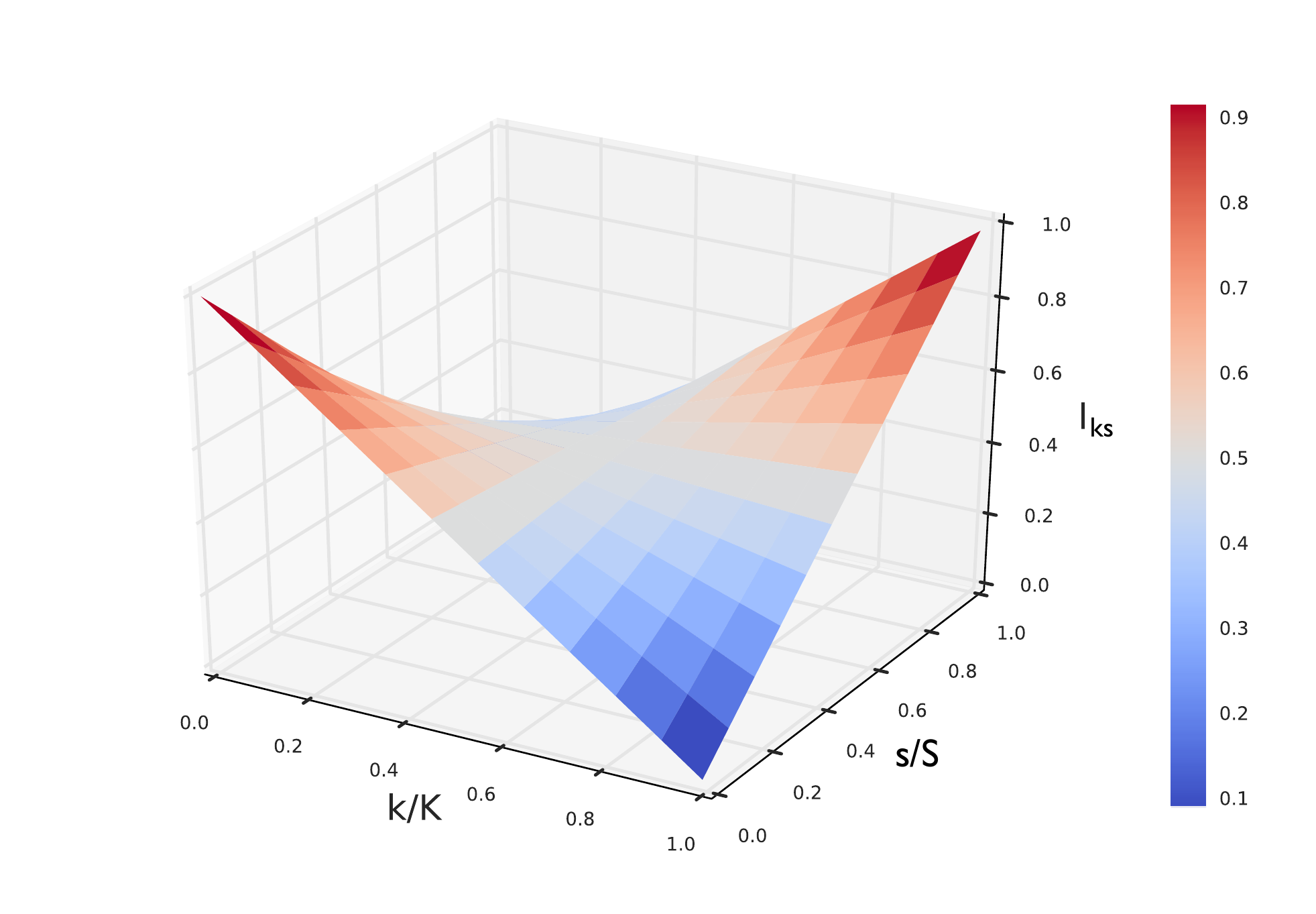}
    \caption{Surface for the position encodings.}
    \label{fig:pe}
\end{figure}

To obtain $l_s$ we first calculate a constant matrix $L$ where we define each element as

\begin{align}\label{eq:pe}
L_{ks} & = (1-k/K)(1-s/\mathcal{S})+\frac{k}{K}\frac{s}{\mathcal{S}},
\end{align}
adapting a formula from ~\cite{Sukhbaatar:2015}. Here, $k\in \{1,2,...,K\}$ is the context vector index and $\mathcal{S}$ is the maximum sequence length across all source sequences. The manifold is shown graphically in Figure \ref{fig:pe}. We can see that earlier encoder states are upweighted in the first context vectors, and later states are upweighted in later vectors. The symmetry of the manifold and its stationary point having value 0.5 both follow from Eq. ~\ref{eq:pe}. The elements of the matrix that fall beyond the sequence lengths are then masked out and the remaining elements are renormalized across the timestep dimension. This results in the jagged array of position encodings $\{l_{ks}\}$.
\section{Experiments}
\label{sec:experiments}

\subsection{Toy Copying Experiment}

Due to the reduction of computational time complexity we expect our method to yield performance gains especially for longer sequences and tasks where the source can be compactly represented in a fixed-size memory matrix. To investigate the trade-off between speed and performance, we compare our technique to standard models with and without attention on a Sequence Copy Task of varying length like in \citet{Graves:2014}. We generated 4 training datasets of 100,000 examples and a validation dataset of 1,000 examples. The vocabulary size was 20. For each dataset, the sequences had lengths randomly chosen between 0 to $L$, for $L\in \{10, 50, 100, 200\}$ unique to each dataset.

\subsubsection{Training Setup}

\begin{figure*}[ht]
\begin{subfigure}[t]{0.49\textwidth}
    \centering
    \includegraphics[trim={30 0 0 0},clip,width=0.95\textwidth]{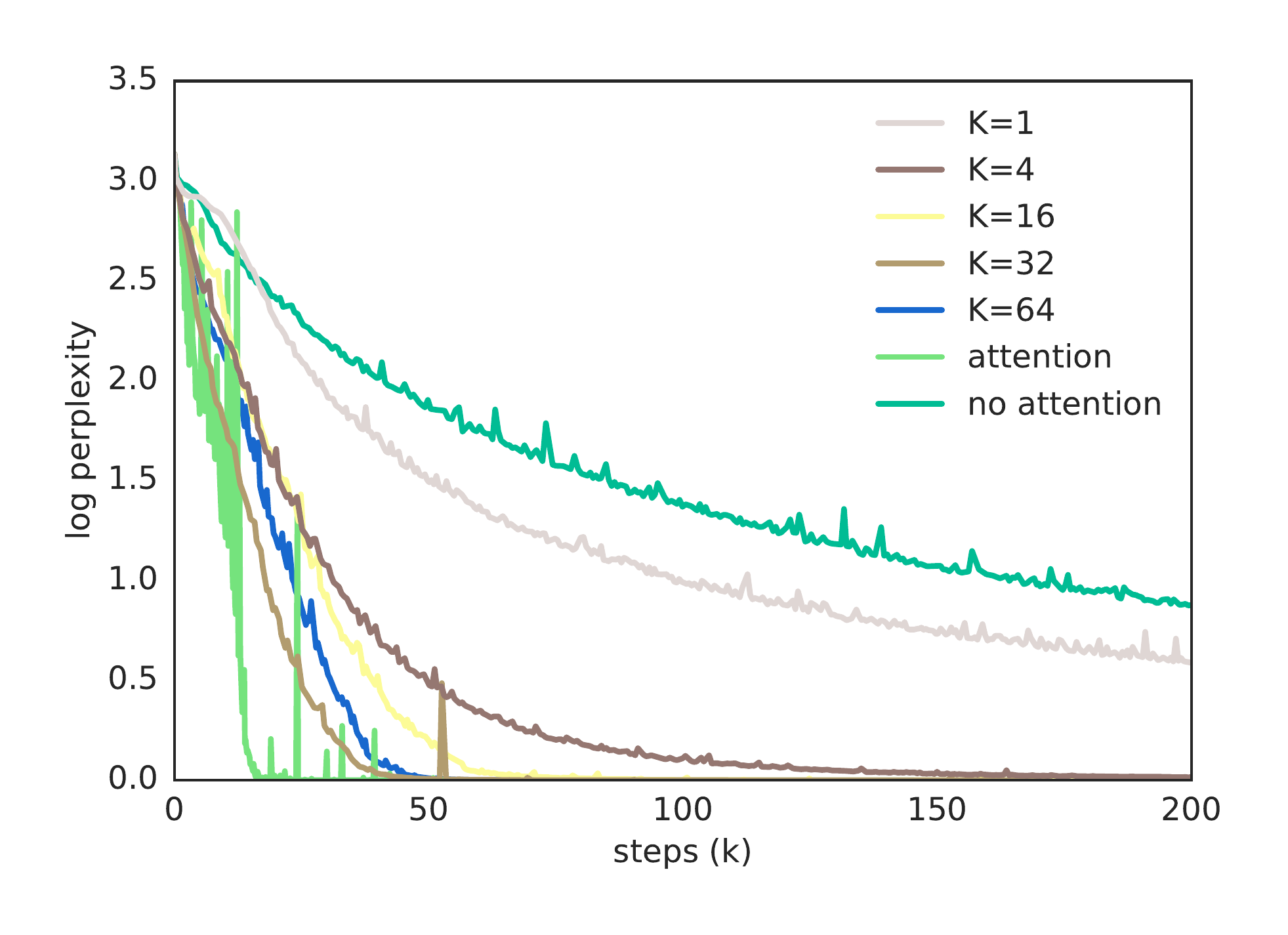}
    \caption{Comparison of varying $K$ for copying sequences of length 200 on evaluation data, showing that large $K$ leads to faster convergence and small $K$ performs similarly to the non-attentional baseline.}
    \label{fig:learning_toy_200}
\end{subfigure}\hfill
\begin{subfigure}[t]{0.49\textwidth}
    \centering
    \includegraphics[trim={0 0 0 0},clip,width=1\textwidth]{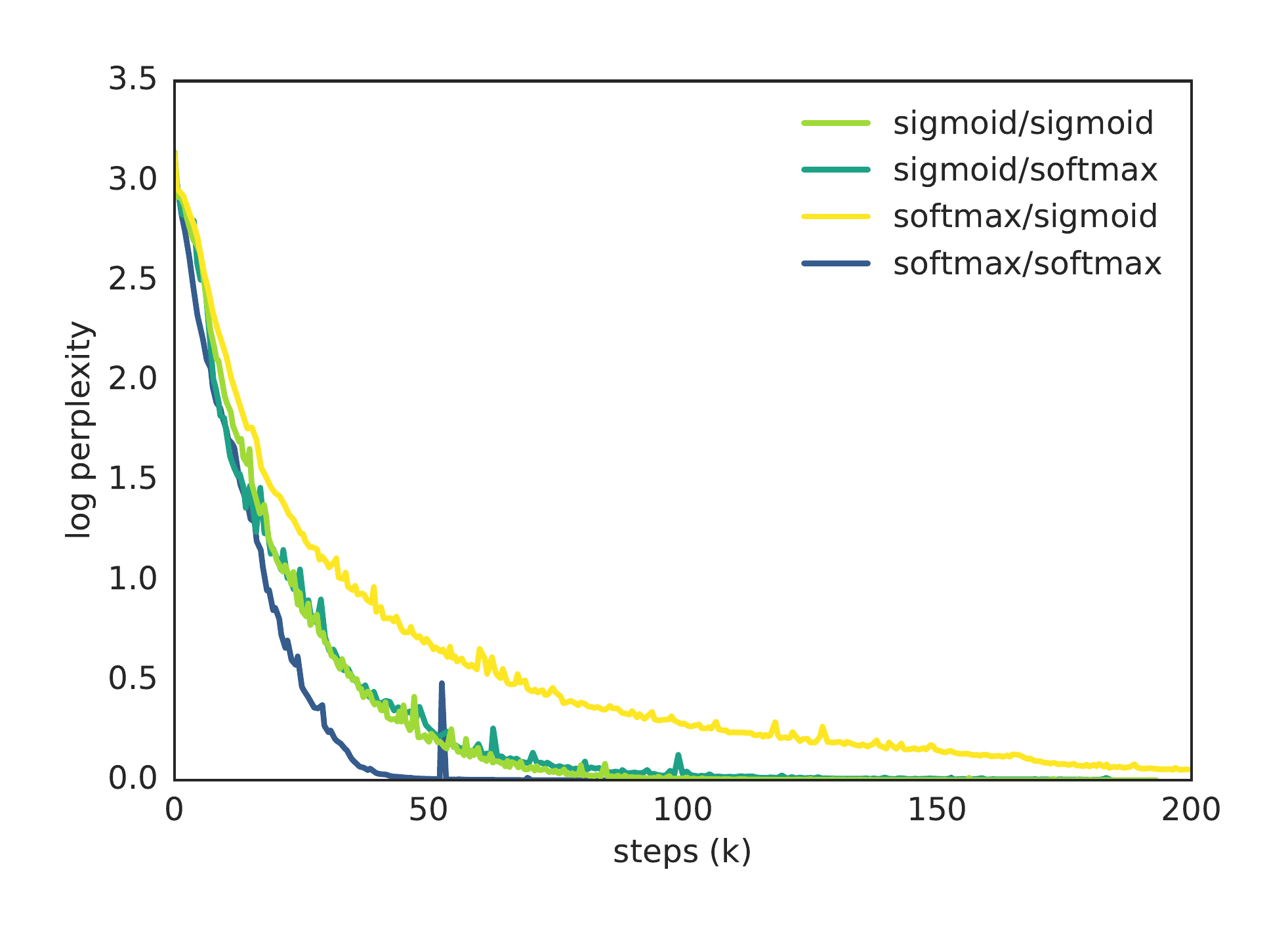}
    \caption{Comparison of sigmoid and softmax functions for choosing the encoder and decoder attention scores on evaluation data, showing that choice of gating/normalization matters.}
    \label{fig:learning_toy_sigsoft}
\end{subfigure}
\caption{Training Curves for the Toy Copy task}
\label{fig:learning_toy}
\end{figure*}

All models are implemented using TensorFlow based on the seq2seq implementation of ~\citet{Britz:2017}\footnote{http://github.com/google/seq2seq} and trained on a single machine with a Nvidia K40m GPU. We use a 2-layer 256-unit, a bidirectional LSTM ~\cite{Schmidhuber:1997} encoder, a 2-layer 256-unit LSTM decoder, and 256-dimensional embeddings. For the attention baseline, we use the standard parametrized attention ~\cite{Bahdanau:2014}. Dropout of 0.2 (0.8 keep probability) is applied to the input of each cell and we optimize using Adam \cite{Kingma:2014} at a learning rate of 0.0001 and batch size 128. We train for at most 200,000 steps (see Figure \ref{fig:learning_toy} for sample learning curves). BLEU scores are calculated on tokenized data using the \textit{multi-bleu.perl} script in Moses.\footnote{http://github.com/moses-smt/mosesdecoder} We decode using beam search with a beam

\begin{table}[tb]
\begin{tabular}{|l|l|r|r|}
    \hline
    \bf Length & \bf Model & \bf BLEU & \bf Time (s) \\
    \hline
     20 & No Att & 99.93 & 2.03 \\
     & $K=1$ & 99.52 & 2.12 \\
     & $K=4$ & 99.56 & 2.25 \\
     & $K=16$ & 99.56 & 2.21 \\
     & $K=32$ & 99.57 & 2.59 \\
     & $K=64$ & 99.75 & 2.86 \\
     & Att & 99.98 & 2.86 \\
     \hline
     50 & No Att & 97.37 & 3.90 \\
     & $K=1$ & 98.86 & 4.33 \\
     & $K=4$ & 99.95 & 4.48 \\
     & $K=16$ & 99.96 & 4.58 \\
     & $K=32$ & 99.96 & 5.35 \\
     & $K=64$ & 99.97& 5.84 \\
     & Att & 99.94 & 6.46 \\
    \hline
     100 & No Att & 73.99 & 6.33 \\    
     & $K=1$ & 87.42 & 7.32 \\
     & $K=4$ & 99.81 & 7.47 \\
     & $K=16$ & 99.97 & 7.50 \\
     & $K=32$ & 99.99 & 7.65 \\
     & $K=64$ & 100.00 & 7.77 \\
     & Att & 100.00 & 11.00 \\
    \hline
     200 & No Att & 32.64 & 9.10 \\
     & $K=1$ & 44.22 & 9.30 \\
     & $K=4$ & 98.54 & 9.49 \\
     & $K=16$ & 99.98 & 9.53 \\
     & $K=32$ & 100.00 & 9.59 \\
     & $K=64$ & 100.00 & 9.78 \\
     & Att & 100.00 & 14.28 \\
    \hline
\end{tabular}
\caption{BLEU scores and computation times with varying $K$ and sequence length compared to baseline models with and without attention.}
\label{table:toy_copy}
\end{table}

\noindent
size of 10 ~\cite{Wiseman:2016}.
 
\subsubsection{Results}

Table \ref{table:toy_copy} shows the  BLEU scores of our model on different sequence lengths while varying $K$. This is a study of the trade-off between computational time and representational power. A large $K$ allows us to compute complex source representations, while a $K$ of 1 limits the source representation to a single vector. We can see that performance consistently increases with $K$ up to a point that depends on the data length, with longer sequences requiring more complex representations. The results with and without position encodings are almost identical on the toy data. Our technique learns to fit the data as well as the standard attention mechanism despite having less representational power. Both beat the non-attention baseline by a significant margin.

That we are able to represent the source sequence with a fixed size matrix with fewer than $|S|$ rows suggests that traditional attention mechanisms may be representing the source with redundancies and wasting computational resources. This makes intuitive sense for the toy task, which should require a relatively simple representation.

The last column shows that our technique significantly speeds up the inference process. The gap in inference speed increases as sequences become longer. We measured inference time on the full validation set of 1,000 examples, not including data loading or model construction times.

Figure \ref{fig:learning_toy_200} shows the learning curves for sequence length 200. We see that $K=1$ is unable to fit the data distribution, while $K\in\{32, 64\}$ fits the data almost as quickly as the attention-based model. Figure \ref{fig:learning_toy_sigsoft} shows the effect of varying the encoder and decoder scoring functions between softmax and sigmoid. All combinations manage to fit the data, but some converge faster than others. In section \ref{sec:visualizing_attention} we show that distinct alignments are learned by different function combinations.


\subsection{Machine Translation}
\label{sec:nmt}

\begin{figure*}[ht]
\centering

\begin{subfigure}[t]{0.44\textwidth}
\centering
\hspace*{+0.23cm}   
\includegraphics[trim={0 150 10 100},clip,width=0.96\textwidth]{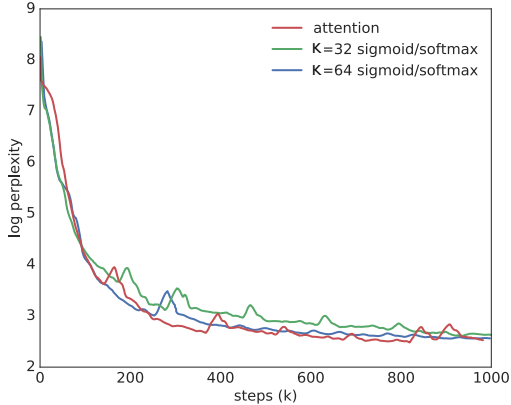}
\caption{Training curves for en-fi}
\label{fig:en_fi_curves}
\end{subfigure}\hfill
\begin{subfigure}[t]{0.432\textwidth}
\centering
\hspace*{-0.785cm}   
\includegraphics[trim={2 150 0 100},clip,width=0.98\textwidth]{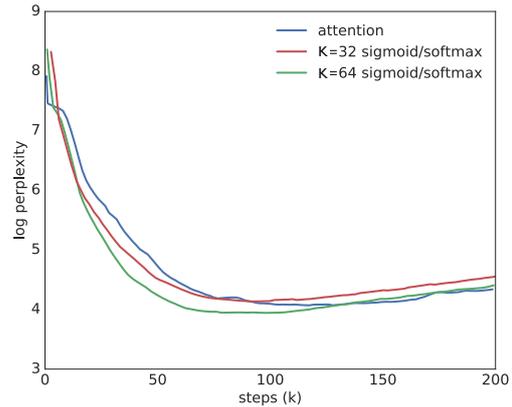}
\caption{Training curves for en-tr}
\label{fig:en_tr_curves}
\end{subfigure}

\caption{Comparing training curves for en-fi and en-tr with sigmoid encoder scoring and softmax decoder scoring and position encoding. Note that en-tr curves converged very quickly.}
\label{fig:nmt_curves}
\end{figure*}

\begin{figure}[h]
\centering\hspace*{-0.45cm}
\includegraphics[width=0.43\textwidth]{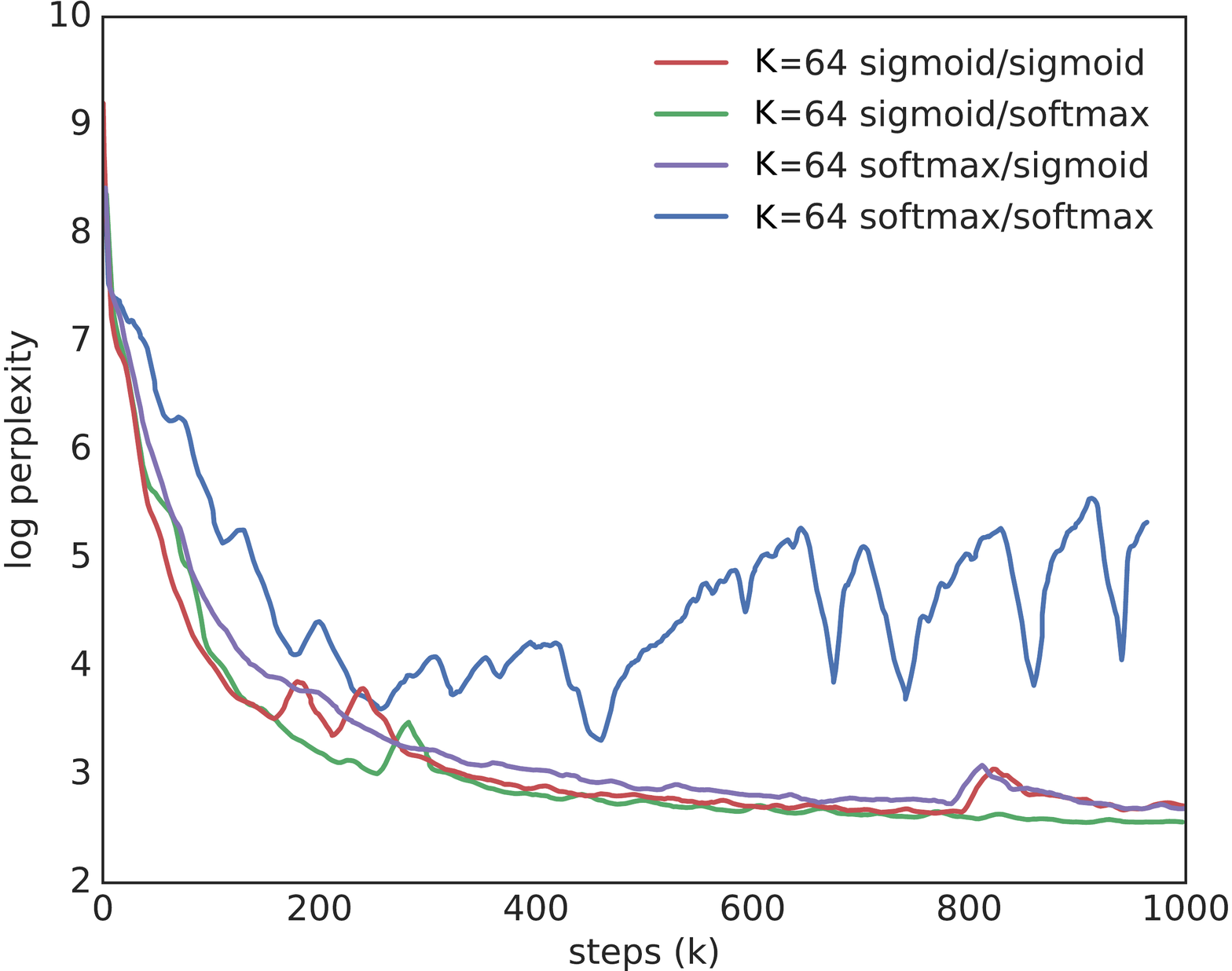}
\caption{Comparing training curves for en-fi for different encoder/decoder scoring functions for our models at $K=64$.}
\label{fig:nmt_gating}
\end{figure}

Next, we explore if the memory-based attention mechanism is able to fit complex  real-world datasets. For this purpose we use 4 large machine translation datasets of WMT'17\footnote{statmt.org/wmt17/translation-task.html} on the following language pairs: English-Czech (en-cs, 52M examples), English-German (en-de, 5.9M examples), English-Finish (en-fi, 2.6M examples), and English-Turkish (en-tr, 207,373 examples). We used the newly available pre-processed datasets for the WMT'17 task.\footnote{http://data.statmt.org/wmt17/translation-task/preprocessed} Note that our scores may not be directly comparable to other work that performs their own data pre-processing. We learn shared vocabularies of 16,000 subword units using the BPE algorithm ~\cite{Sennrich:2015}. We use newstest2015 as a validation set, and report BLEU on newstest2016.
 
\subsubsection{Training Setup}

We use a similar setup to the Toy Copy task, but use 512 RNN and embedding units, train using 8 distributed workers with 1 GPU each, and train for at most 1M steps. We save checkpoints every 30 minutes during training, and choose the best based on the validation BLEU score.

\subsubsection{Results}

\begin{table*}[t]
\begin{centering}
\begin{tabular}{|l|l|l|l|c|c|c|c|c|}
    \hline
    Model & Dataset & $\emph{K}$         & en-cs & en-de & en-fi & en-tr \\
    \hline
    Memory Attention & Test & 32       & 19.37 & 28.82 & 15.87 & - \\
    && 64                       & 19.65 & 29.53 & 16.49 & - \\
    & Valid & 32                  & 19.20 & 26.20 & 15.90 & 12.94 \\
    && 64                       & 19.63 & 26.39 & 16.35 & 13.06\\
    \hline
    Memory Attention + PE & Test & 32    & 19.45 & 29.53 & 15.86 & - \\
    && 64                       & \textbf{20.36} & 30.61 & 17.03 & - \\
    & Valid & 32                  & 19.35 & 26.22 & 16.31 & 12.97\\
    && 64                       & 19.73 & 27.31 & 16.91 & 13.25 \\
    \hline
    Attention & Test & -         & 19.19 & \textbf{30.99} & \textbf{17.34} & - \\
    & Valid & -                    & 18.61 & 28.13 & 17.16 & 13.76 \\
    \hline
\end{tabular}
\caption{BLEU scores on WMT'17 translation datasets from the memory attention models and regular attention baselines. We picked the best out of the four scoring function combinations on the validation set. Note that en-tr does not have an official test set. Best test scores on each dataset are highlighted.}
\label{table:nmt}
\end{centering}
\end{table*}

\begin{table}[t]
\centering
\begin{tabular}{|l|c|}
\hline
\bf Model & \bf Decoding Time (s) \\
\hline
$K=32$ & 26.85 \\
$K=64$ & 27.13 \\
Attention & 33.28 \\
\hline
\end{tabular}
\caption{Decoding time, averaged across 10 runs, for the en-de validation set (2169 examples) with average sequence length of 35. Results are similar for both PE and non-PE models. }
\label{table:nmt_decoding_time}
\end{table}

Table \ref{table:nmt} compares our approach with and without position encodings, and with varying values for hyperparameter $K$, to baseline models with regular attention mechanism. Learning curves are shown in Figure \ref{fig:nmt_curves}. We see that our memory attention model with sufficiently high $K$ performs on-par with, or slightly better, than the attention-based baseline model despite its simpler nature. Across the board, models with $K=64$ performed better than corresponding models with $K=32$, suggesting that using a larger number of attention vectors can capture a richer understanding of source sequences. Position encodings also seem to consistently improve model performance.

Table \ref{table:nmt_decoding_time} shows that our model results in faster decoding time even on a complex dataset with a large vocabulary of 16k. We measured decoding time over the full validation set, not including time used for model setup and data loading, averaged across 10 runs. The average sequence length for examples in this data was 35, and we expect more significant speedups for tasks with longer sequences, as suggested by our experiments on toy data. Note that in our NMT examples/experiments, $K\approx T$, but we obtain computational savings from the fact that $K \ll D$. We may be able to set $K \ll T$, as in toy copying, and still get very good performance in other tasks. For instance, in summarization the source is complex but the representation of the source required to perform the task is "simple" (i.e. all that is needed to generate the abstract). 

Figure \ref{fig:nmt_gating} shows the effect of using sigmoid and softmax function in the encoders and decoders. We found that softmax/softmax consistently performs badly, while all other combinations perform about equally well. We report results for the best combination only (as chosen on the validation set), but we found this choice to only make a minor difference.

\section{Visualizing Attention}
\label{sec:visualizing_attention}

\begin{figure*}[htb!]
\centering
\begin{subfigure}[b]{0.475\textwidth}
    \includegraphics[width=\textwidth]{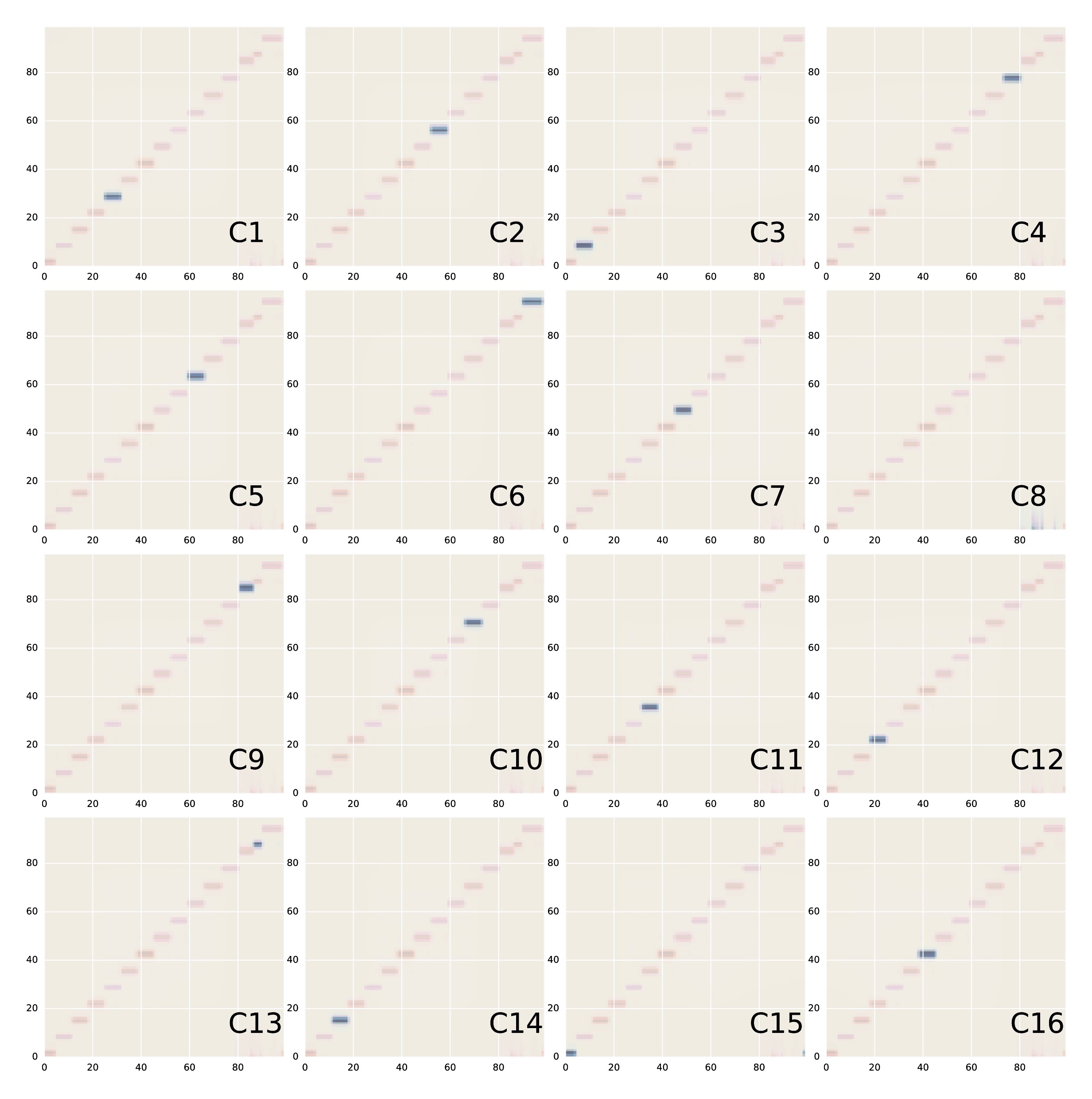}
\end{subfigure}\hspace{.11 cm}
\begin{subfigure}[b]{0.475\textwidth}
    \includegraphics[width=\textwidth]{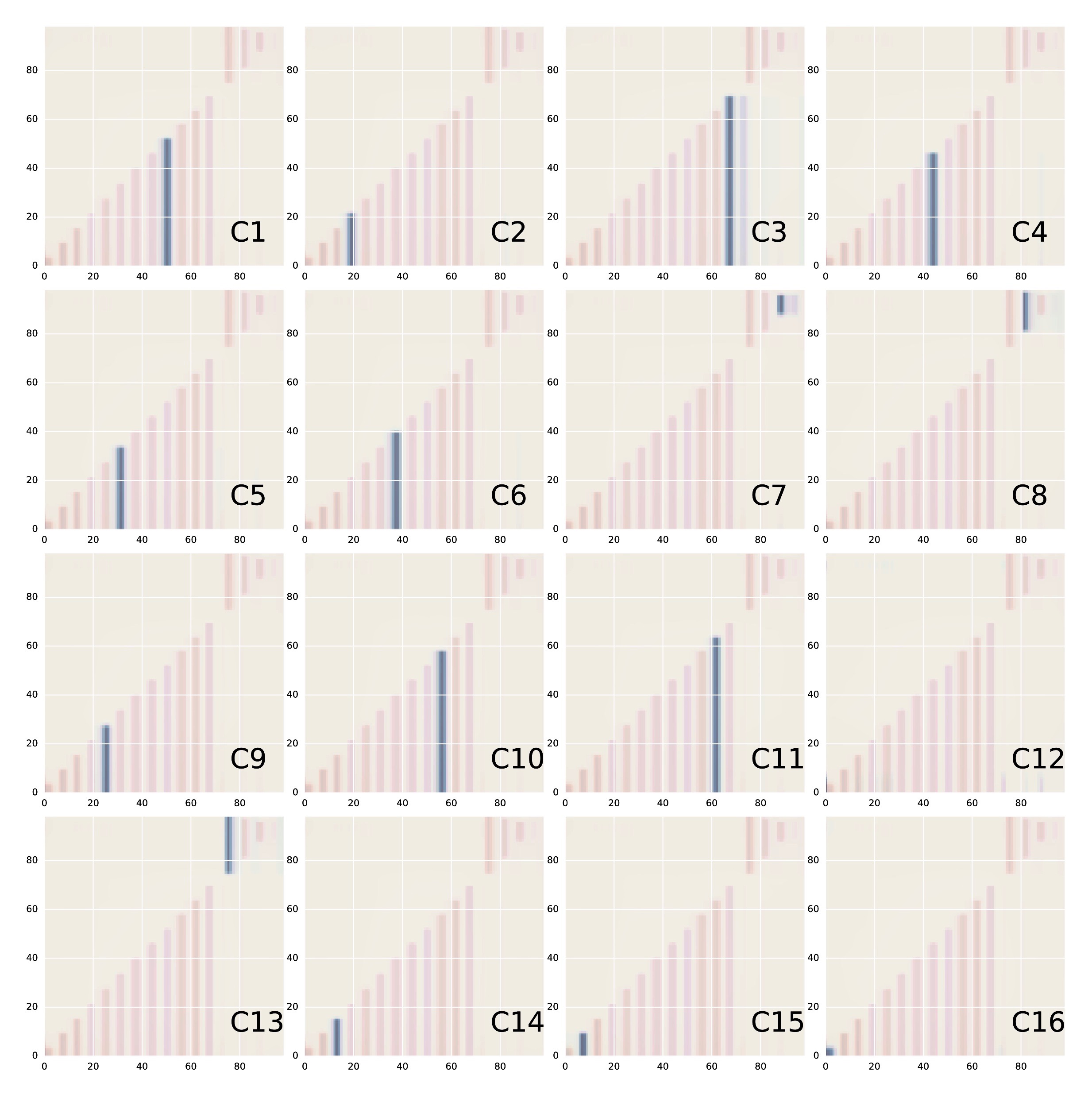}
\end{subfigure}\\
\begin{subfigure}[b]{1.\textwidth}
\includegraphics[trim={0 85cm 0 85cm},clip,width=1.\textwidth]{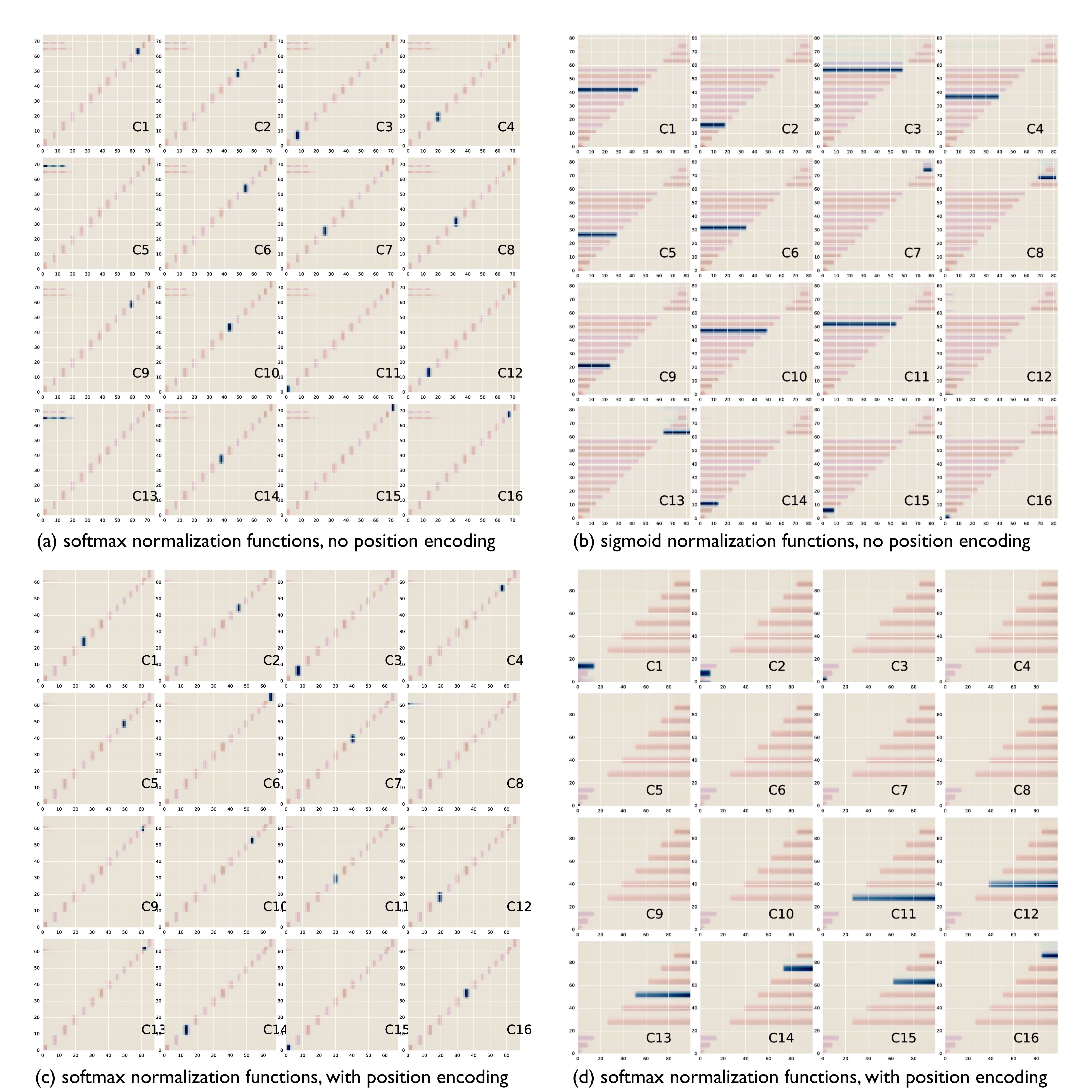}
\end{subfigure}
\caption{Attention scores at each step of decoding for on a sample from the sequence length 100 toy copy dataset. Individual attention vectors are highlighted in blue. ($y$-axis: source tokens; $x$-axis: target tokens)}
\label{fig:k16_attention}
\end{figure*}
\begin{figure*}[htb!]
\vspace{-0.14 cm}
\centering
\begin{subfigure}[b]{0.385\textwidth}
    \includegraphics[trim={3 100 0 145},clip,width=\textwidth]{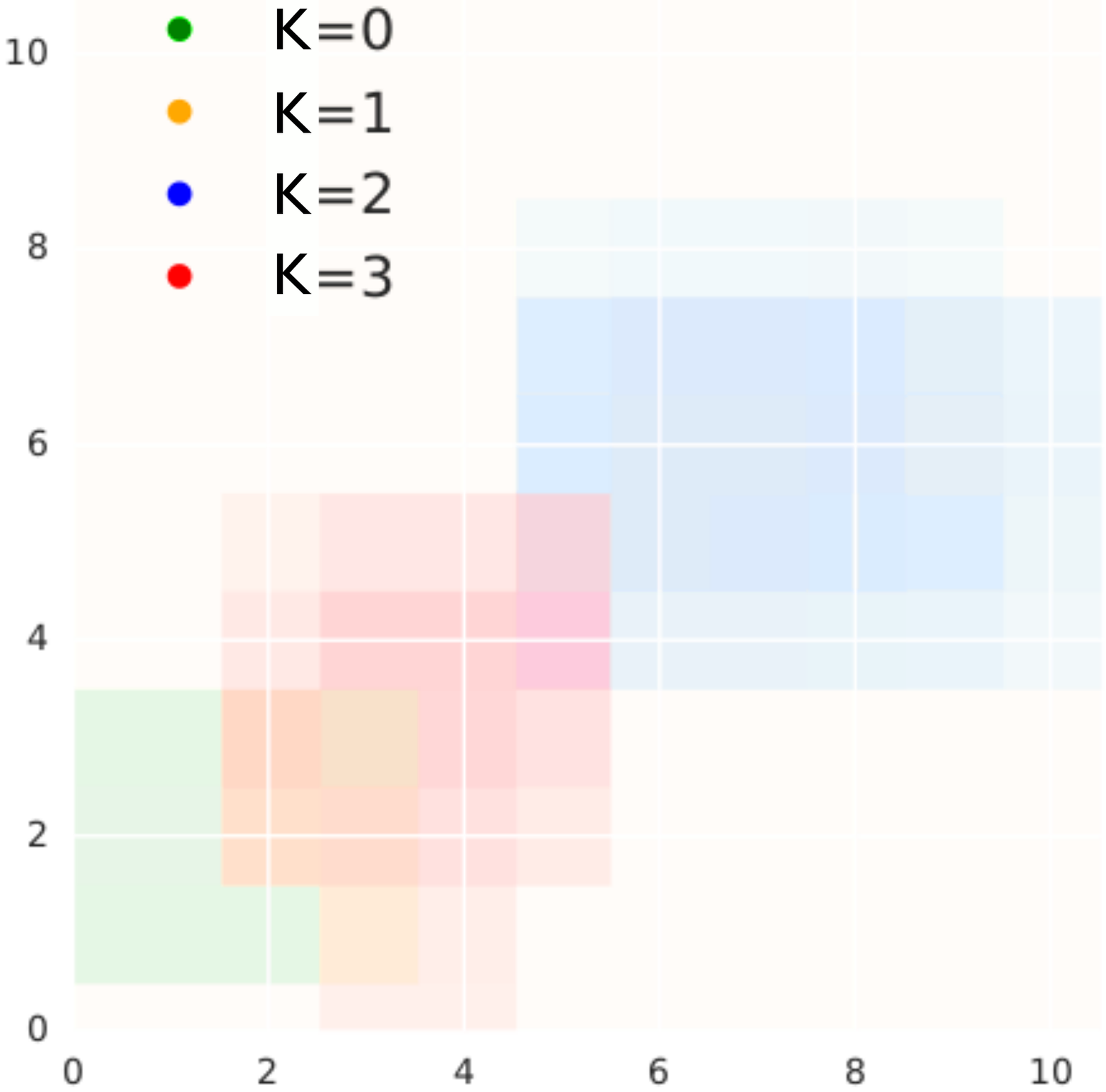}
\end{subfigure}
\begin{subfigure}[b]{0.375\textwidth}
    \includegraphics[trim={0 0 0 45},clip,width=\textwidth]{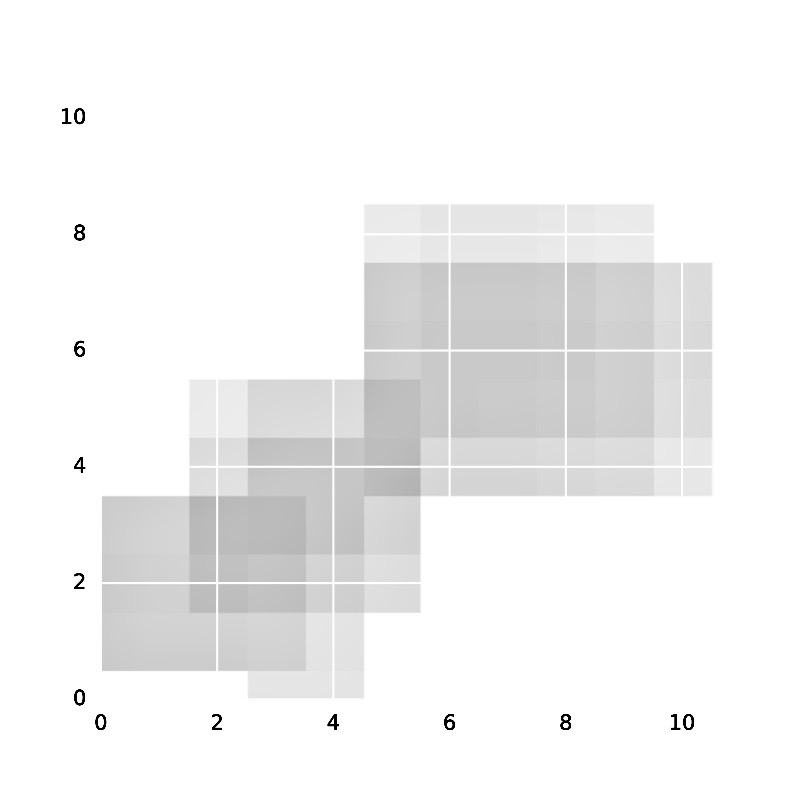}
\end{subfigure}
\caption{Attention scores at each step of decoding for $K=4$ on a sample with sequence length 11. The subfigure on the left color codes each individual attention vector. ($y$-axis: source; $x$-axis: target)}
\label{fig:k4_attention}
\end{figure*}
\begin{figure*}[htb!]
\vspace{-0.08 cm}
\centering
\begin{subfigure}[b]{0.73\textwidth}
    $\vcenter{\hbox{\includegraphics[width=\textwidth]{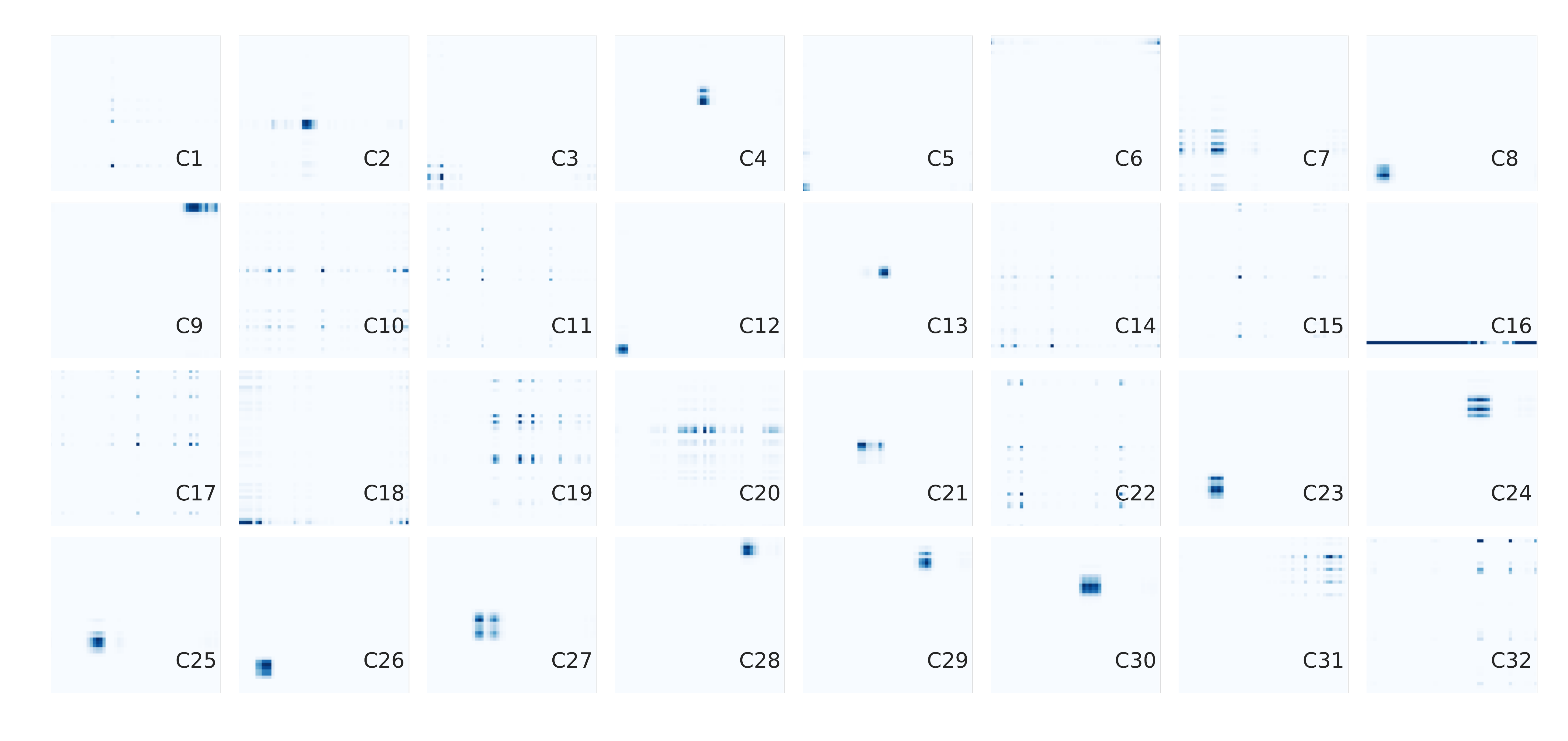}}}$
\end{subfigure}
\begin{subfigure}[b]{0.23\textwidth}
    $\vcenter{\hbox{\includegraphics[width=\textwidth]{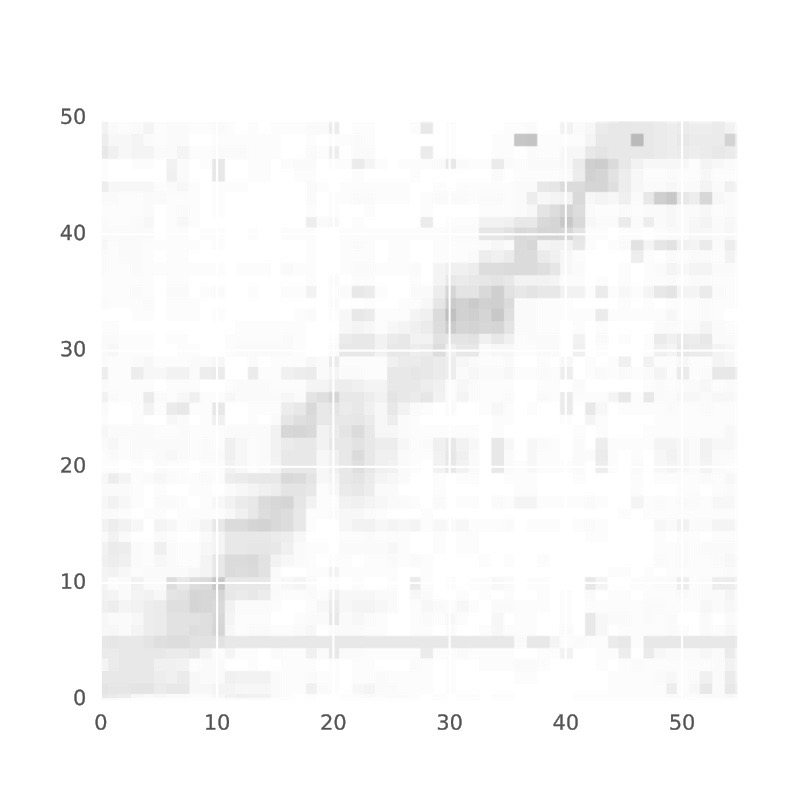}}}$
\end{subfigure}
\caption{Attention scores at each step of decoding for en-de WMT translation task using model with sigmoid scoring functions and $K=32$. The left subfigure displays each individual attention vector separately while the right subfigure displays the full combined attention. ($y$-axis: source; $x$-axis: target)}
\label{fig:attention_scores}
\end{figure*}

A useful property of the standard attention mechanism is that it produces meaningful alignment between source and target sequences. Often, the attention mechanism learns to progressively focus on the next source token as it decodes the target. These visualizations can be an important tool in debugging and evaluating seq2seq models and are often used for unknown token replacement.

This raises the question of whether or not our proposed memory attention mechanism also learns to generate meaningful alignments. Due to limiting the number of attention contexts to a number that is generally less than the sequence length, it is not immediately obvious what each context would learn to focus on. Our hope was that the model would learn to focus on multiple alignments at the same time, within the same attention vector. For example, if the source sequence is of length 40 and we have $K=10$ attention contexts, we would hope that $C_1$ roughly focuses on tokens 1 to 4, $C_2$ on tokens 5 to 8, and so on. Figures \ref{fig:k16_attention} and \ref{fig:k4_attention} show that this is indeed the case. To generate this visualization we multiply the attention scores $\alpha$ and $\beta$ from the encoder and decoder. Figure \ref{fig:attention_scores} shows a sample translation task visualization.

Figure \ref{fig:k16_attention} suggests that our model learns distinct ways to use its memory depending on the encoder and decoder functions. Interestingly, using softmax normalization results in attention maps typical of those derived from using standard attention, i.e. a relatively linear mapping between source and target tokens. Meanwhile, using sigmoid gating results in what seems to be a distributed representation of the source sequences across encoder time steps, with multiple contiguous attention contexts being accessed at each decoding step.



\section{Related Work}

Our contributions build on previous work in making seq2seq models more computationally efficient. ~\citet{Luong:2015} introduce various attention mechanisms that are computationally simpler and perform as well or better than the original one presented in \citet{Bahdanau:2014}. However, these typically still require $O(D^2)$ computation complexity, or lack the flexibility to look at the full source sequence. Efficient location-based attention ~\cite{Xu:2015} has also been explored in the image recognition domain.

~\citet{Wu:2016} presents several enhancements to the standard seq2seq architecture that allow more efficient computation on GPUs, such as only attending on the bottom layer. ~\citet{Kalchbrenner:2016} propose a linear time architecture based on stacked convolutional neural  networks. ~\citet{Gehring:2016} also propose the use of convolutional encoders to speed up NMT. ~\citet{Brebisson:2016} propose a linear attention mechanism based on covariance matrices applied to information retrieval. ~\citet{Raffel:2017} enable online linear time attention calculation by enforcing that the alignment between input and output sequence elements be monotonic. Previously, monotonic attention was proposed for morphological inflection generation by ~\citet{aharonimorphological}.


\section{Conclusion}
In this work, we propose a novel memory-based attention mechanism that results in a linear computational time of $O(KD(|S| + |T|))$ during decoding in seq2seq models. Through a series of experiments, we demonstrate that our technique leads to consistent inference speedups as sequences get longer, and can fit complex data distributions such as those found in Neural Machine Translation. We show that our attention mechanism learns meaningful alignments despite being constrained to a fixed representation after encoding. We encourage future work that explores the optimal values of $K$ for various language tasks and examines whether or not it is possible to predict $K$ based on the task at hand. We also encourage evaluating our models on other tasks that must deal with long sequences but have compact representations, such as summarization and question-answering, and further exploration of their effect on memory and training speed.


\vfill

\bibliography{emnlp2017}
\bibliographystyle{emnlp_natbib}

\appendix

\end{document}